\pdfoutput=1

\documentclass[11pt]{article}

\usepackage[]{emnlp2021}

\usepackage{times}
\usepackage{latexsym}
\usepackage{todonotes}
\usepackage{enumitem}
\usepackage{lipsum}
\usepackage{booktabs}
\usepackage{array}
\usepackage{multirow}
\usepackage{graphicx}
\usepackage[normalem]{ulem}

\usepackage{etoolbox}
\usepackage{todonotes}
\usepackage{hyperref}
\usepackage{enumerate}
\usepackage{enumitem}
\usepackage{amsmath}
\usepackage{color}
\usepackage{tcolorbox}
\usepackage{CJK}
\usepackage{adjustbox}
\usepackage{caption}
\usepackage{subcaption}
\tcbset{width=0.9\textwidth,boxrule=0pt,colback=red,arc=0pt,auto outer arc,left=0pt,right=0pt,boxsep=5pt}
\usepackage[font=small,skip=-5pt]{subcaption}
\usepackage{diagbox}

\usepackage[T1]{fontenc}

\usepackage[utf8]{inputenc}

\usepackage{microtype}

\newcommand\METHOD{\textsc{MT-PET}}

\newcommand\NTEST{\textsc{563}}

\title{Semi-Supervised Exaggeration Detection of Health Science Press Releases}

\author{Dustin Wright \and Isabelle Augenstein \\
  Dept. of Computer Science \\
  University of Copenhagen \\
  Denmark \\
  \texttt{\{dw|augenstein\}@di.ku.dk}}

\begin{document}
\maketitle
\begin{abstract}
Public trust in science depends on honest and factual communication of scientific papers. However, recent studies have demonstrated a tendency of news media to misrepresent scientific papers by exaggerating their findings. Given this, we present a formalization of and study into the problem of \textit{exaggeration detection} in science communication. 
While there are an abundance of scientific papers and popular media articles written about them, very rarely do the articles include a direct link to the original paper, making data collection challenging. We address this by curating a set of labeled press release/abstract pairs from existing expert annotated studies on exaggeration in press releases of scientific papers suitable for benchmarking the performance of machine learning models on the task. Using limited data from this and previous studies on exaggeration detection in science, we introduce \METHOD, a multi-task version of Pattern Exploiting Training (PET), which leverages knowledge from complementary cloze-style QA tasks to improve few-shot learning. We demonstrate that \METHOD\ outperforms PET and supervised learning both when data is limited, as well as when there is an abundance of data for the main task.\footnote{The code and data are available online at \url{https://github.com/copenlu/scientific-exaggeration-detection}}


\end{abstract}

\section{Introduction}
\begin{figure}[t]
  
  \centering
    \includegraphics[width=\linewidth]{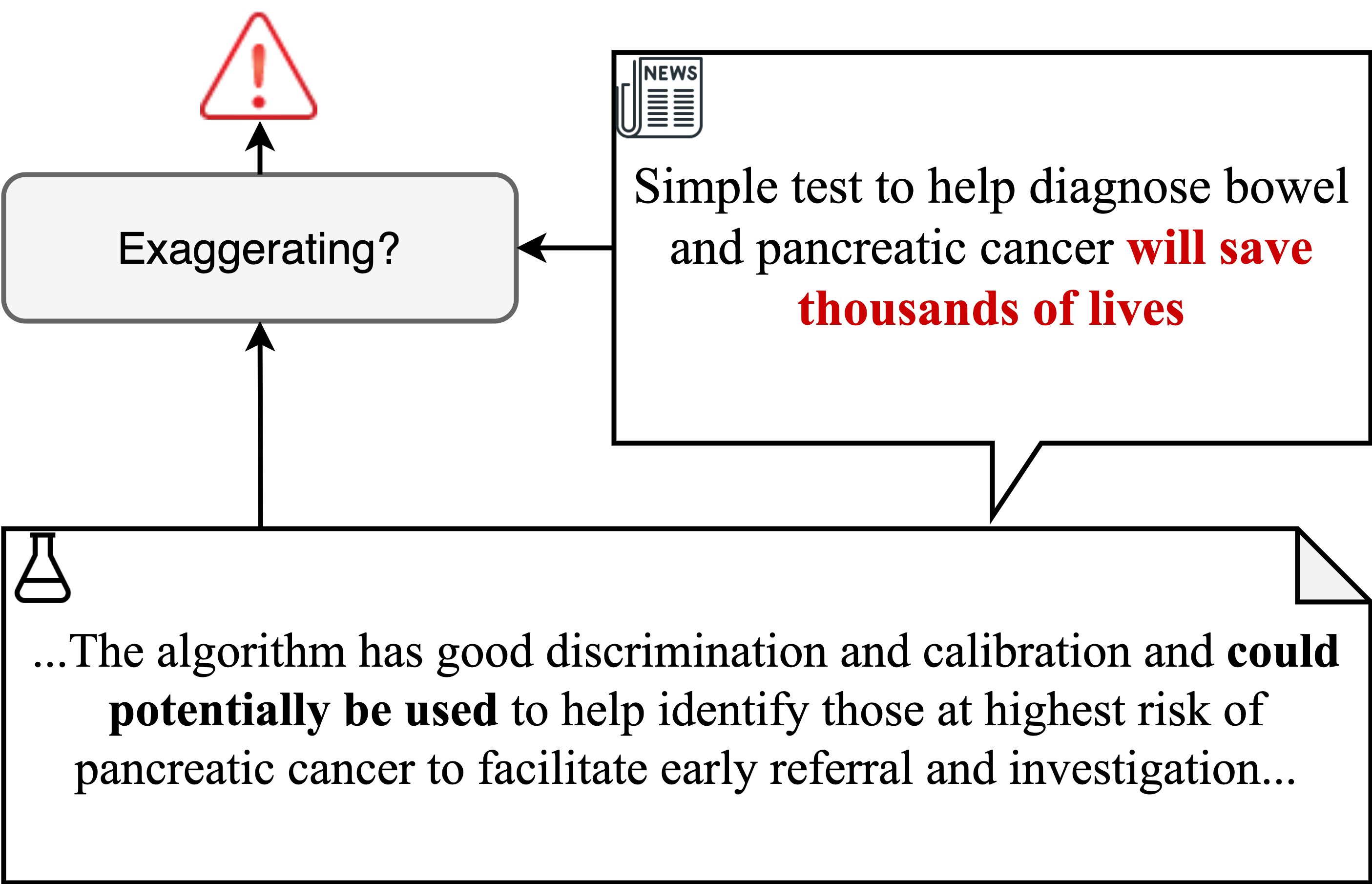}
    \caption{Scientific exaggeration detection is the problem of identifying when a news article reporting on a scientific finding has exaggerated the claims made in the original paper. In this work, we are concerned with predicting exaggeration of the main finding of a scientific abstract as reported by a press release.}
    \label{fig:exaggeration-detection}
\end{figure}
Factual and honest science communication is important for maintaining public trust in science~\cite{nelkin1987selling,moore2006bad}, and the ``dominant link between academia and the media'' are press releases about scientific articles~\cite{sumner2014association}. However, multiple studies have demonstrated that press releases have a significant tendency to sensationalize their associated scientific articles \cite{sumner2014association,bratton2019association,woloshin2009press,woloshin2002press}. 
In this paper, we explore how natural language processing can help identify exaggerations of scientific papers in press releases.

While \citet{sumner2014association} and \citet{bratton2019association} performed manual analyses to understand the prevalence of exaggeration in press releases of scientific papers from a variety of sources, recent work has attempted to expand this using methods from NLP \cite{yu2019detecting,yu2020measuring,DBLP:conf/emnlp/LiZY17}. These works focus on the problem of automatically detecting the difference in the strength of causal claims made in scientific articles and press releases. They accomplish this by first building datasets of main claims taken from PubMed abstracts and (unrelated) press releases from EurekAlert\footnote{\url{https://www.eurekalert.org/}} labeled for their strength. With this, they train machine learning models to predict claim strength, and analyze unlabelled data using these models. This marks an important first step toward the goal of automatically identifying exaggerated scientific claims in science reporting.

However, existing work has only partially attempted to address this task using NLP. Particularly, there exists no standard benchmark data for the exaggeration detection task with \textbf{paired} press releases and abstracts i.e. where the data consist of tuples of the form (press release, abstract) and the press release is written about the paired scientific paper. Collecting paired data labeled for exaggeration is critical for understanding how well any solution performs on the task, but is challenging and expensive as it requires domain expertise~\cite{sumner2014association}. The focus of this work is then to curate a standard set of benchmark data for the task of scientific exaggeration detection, provide a more realistic task formulation of the problem, and develop methods effective for solving it using limited labeled data. To this end, we present~\METHOD, a multi-task implementation of Pattern Exploiting Training (PET, \citet{schick2020exploiting,schick2020small}) for detecting exaggeration in health science press releases. We test our method by curating a benchmark test set of data from the expert annotated data of \citet{sumner2014association} and \citet{bratton2019association}, which we release to help advance research on scientific exaggeration detection.

\paragraph{Contributions}
In sum, we introduce:
\vspace{-0.3cm}
\begin{itemize}[noitemsep]
    \item A new, more realistic task formulation for scientific exaggeration detection.
    \item A curated set of benchmark data for testing methods for scientific exaggeration detection consisting of \NTEST~press release/abstract pairs.
    \item \METHOD, a multi-task extension of PET which beats strong baselines on scientific exaggeration detection.
\end{itemize}

\section{Problem Formulation}
We first provide a formal definition of the problem of scientific exaggeration detection, which guides the approach described in \S\ref{sec:approach}. We start with a set of document pairs $\{(t,s) \in \mathcal{D}\}$, where $s$ is a source document (e.g. a scientific paper abstract) and $t$ is a document written about the source document $s$ (e.g. a press release for the paper). The goal is to predict a label $l \in \{0,1,2\}$ for a given document pair $(t,s)$, where $0$ implies the target document \textit{undersells} source document, $1$ implies the target document accurately reflects the source document, and $2$ implies the target document \textit{exaggerates} the source document. 

Two realizations of this formulation are investigated in this work. The first (defined as \textbf{T1}) is an \textit{inference} task consisting of labeled document pairs used to learn to predict $l$ directly. In other words, we are given training data of the form $(t, s, l)$ and can directly train a model to predict $l$ from both $t$ and $s$. The second (defined as \textbf{T2}) is as a \textit{classification} task consisting of a training set of documents $d \in \mathcal{D}'$ from \textbf{both} the source and the target domain, and a classifier is trained to predict the \textit{claim strength} $l'$ of sentences from these documents. In other words, we don't require \textbf{paired} documents $(t,s)$ at train time. At test time, these classifiers are then applied to document pairs $(t,s)$ and the predicted claim strengths $(l'_{s}, l'_{t})$ are compared to get the final label $l$. Previous work has used this formulation to estimate the prevalence of \textit{correlation to causation} exaggeration in press releases~\cite{yu2020measuring}, but have not evaluated this on paired labeled instances.

 Following previous work~\cite{yu2020measuring}, we simplify the problem by focusing on detecting when the \textit{main finding} of a paper is exaggerated. The first step is then to identify the main finding from $s$, and the sentence describing the main finding in $s$ from $t$. In our semi-supervised approach, we do this as an intermediate step to acquire unlabeled data, but for all labeled training and test data, we assume the sentences are already identified and evaluate on the sentence-level exaggeration detection task.

\section{Approach}
\label{sec:approach}
\begin{figure}[t]
  
  \centering
    \includegraphics[width=\linewidth]{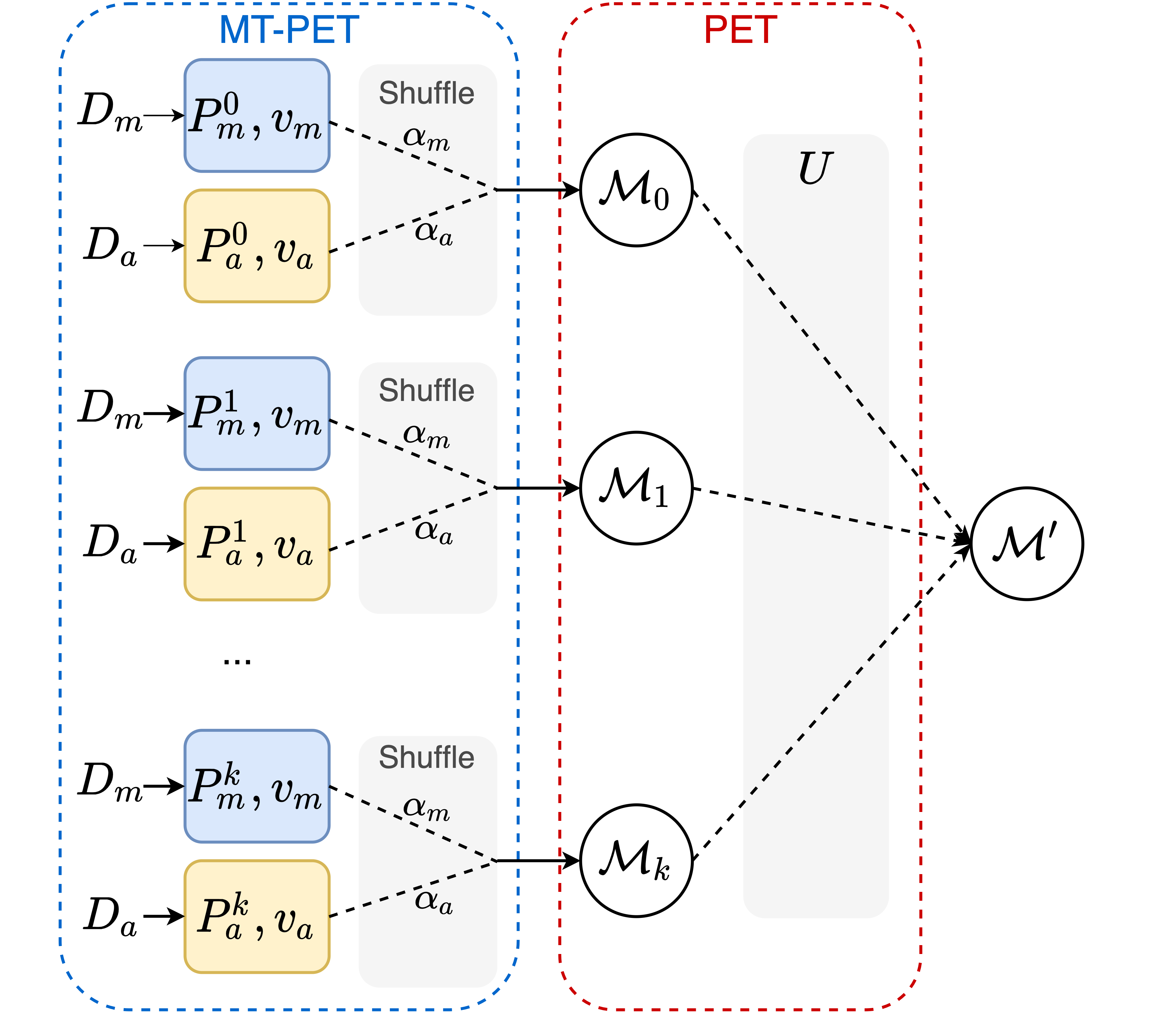}
    \caption{MT-PET design. We define pairs of complementary pattern-verbalizer pairs for a main task and auxiliary task. These PVPs are then used to train PET on data from both tasks.}
    \label{fig:mt-pet}
\end{figure}
One of the primary challenges for scientific exaggeration detection is a lack of labeled training data. Given this, we develop a semi-supervised approach for few-shot exaggeration detection based on pattern exploiting training (PET, \citet{schick2020exploiting,schick2020small}). Our method, multi-task PET (\METHOD, see \autoref{fig:mt-pet}), improves on PET by using multiple complementary cloze-style QA tasks derived from different source tasks during training. 
We first describe PET, followed by \METHOD.

\subsection{Pattern Exploiting Training (PET)}
PET~\cite{schick2020exploiting} uses the masked language modeling objective of pretrained language models to transform a task into one or more cloze-style question answering tasks. 
The two primary components of PET are \textit{patterns} and \textit{verbalizers}. \textit{Patterns} are cloze-style sentences which mask a single token e.g. in sentiment classification with the sentence ``We liked the dinner'' a possible pattern is: ``We liked the dinner. It was \texttt{[MASK]}.'' \textit{Verbalizers} are single tokens which capture the meaning of the task's labels in natural language, and which the model should predict to fill in the masked slots in the provided patterns (e.g. in the sentiment analysis example, the verbalizer could be \texttt{Good}). 

Given a set of \textit{pattern-verbalizer pairs (PVPs)}, an ensemble of models is trained on a small labeled seed dataset to predict the appropriate verbalizations of the labels in the masked slots. These models are then applied on unlabeled data, and the raw logits are combined as a weighted average to provide soft-labels for the unlabeled data. A final classifier is then trained on the soft labeled data using a distillation loss based on KL-divergence.

\subsection{Notation}
We adopt the notation in the original PET paper~\cite{schick2020exploiting} to describe \METHOD. In this, we have a masked language model $\mathcal{M}$ with a vocabulary $V$ and mask token \texttt{[MASK]} $\in V$. A pattern is defined as a function $P(x)$ which transforms a sequence of input sentences $\mathbf{x} = (s_{0},...,s_{k-1}), s_{i} \in V^{*}$ to a phrase or sentence which contains exactly one mask token. Verbalizers $v(x)$ map a label in the task's label space $\mathcal{L}$ to a set of tokens in the vocabulary $V$ which $\mathcal{M}$ is trained to predict.

For a given sample $\mathbf{z} \in V^{*}$ containing exactly one mask token and $w \in V$ corresponding to a word in the language model's vocabulary, $M(w|\mathbf{z})$ is defined as the unnormalized score that the language model gives to word $w$ at the masked position in $\mathbf{z}$. The score for a particular label as given in \citet{schick2020exploiting} is then
\begin{equation}
    s_{\mathbf{p}}(l|\mathbf{x}) = M(v(l) | P(\mathbf{x}))
\end{equation}
For a given sample, PET then assigns a score $s$ for each label based on all of the verbalizations of that label. When applied to unlabeled data, this produces soft labels from which a final model $\mathcal{M}'$ can be trained via distillation using KL-divergence.

\subsection{\METHOD}
\begin{table*}
    \def\arraystretch{1.4}
        \centering
        \fontsize{10}{10}\selectfont
        \begin{tabular}{c c}
        \toprule 
        Name & Pattern\\
        \midrule 
    $P_{T_{1}}^{0}(x)$&Scientists claim $a$. || Reporters claim $b$.The reporters claims are \texttt{[MASK]} \\
    $P_{T_{1}}^{1}(x)$& Academic literature claims $a$. || Popular media claims $b$. The media claims are \texttt{[MASK]} \\
    \midrule
    $P_{T_{2}}^{0}(x)$& [Reporters|Scientists] say $a$. The claim strength is \texttt{[MASK]} \\
    $P_{T_{2}}^{1}(x)$&[Academic literature|Popular media] says $a$. The claim strength is \texttt{[MASK]}\\
    \bottomrule
    \end{tabular}
    \caption{Patterns for both \textbf{T1} (exaggeration detection) and \textbf{T2} (claim strength prediction)}
    \label{tab:patterns}
\end{table*}

In the original PET implementation, PVPs are defined for a single target task. \METHOD~extends this by allowing for auxiliary PVPs from related tasks, adding complementary cloze-style QA tasks during training. The motivation for the multi-task approach is two-fold: 1) complementary cloze-style tasks can potentially help the model to learn different aspects of the main task; in our case, the similar tasks of exaggeration detection and claim strength prediction; 2) data on related tasks can be utilized during training, which is important in situations where data for the main task is limited.

Concretely, we start with a main task $T_{m}$ with a small labeled dataset $(x_{m},y_{m}) \in D_{m}$, where $y_{m} \in \mathcal{L}_{m}$ is a label for the instance, as well as an auxiliary task $T_{a}$ with labeled data $(x_{a},y_{a}) \in D_{a}, y_{a} \in \mathcal{L}_{a}$. Each pattern $P_{m}^{i}(x)$ for the main task has a corresponding complementary pattern $P_{a}^{i}(x)$ for the auxiliary task. Additionally, the labels in $\mathcal{L}_{a}$ have their own verbalizers $v_{a}(x)$. Thus, with $k$ patterns, the full set of PVP tuples is given as 
\begin{equation*}
    \mathcal{P} = \{((P_{m}^{i}, v_{m}), (P_{a}^{i}, v_{a})) | 0 \le i < k\}
\end{equation*}
Finally, a large set of unlabeled data $U$ for the \textit{main task only} is available. \METHOD~then trains the ensemble of $k$ masked language models using the pairs defined for the main and auxiliary task. In other words, for each individual model both the main PVP $(P_{m},v_{m})$ and auxiliary PVP $(P_{a},v_{a})$ are used during training. 

For a given model $\mathcal{M}_{i}$ in the ensemble, on each batch we randomly select one task $T_{c}, c \in \{m,a\}$ on which to train. The PVP for that task is then selected as $(P_{c}^{i}, v_{c})$. Inputs $(x_{c}, y_{c})$ from that dataset are passed through the model, producing raw scores for each label in the task's label space.
\begin{equation}
    s_{\mathbf{p}_{c}^{i}}(\cdot|\mathbf{x}_{c}) = \{\mathcal{M}_{i}(v_{c}(l)|P_{c}^{i}(\mathbf{x}_{c})) | \forall~ l \in \mathcal{L}_{c}\}
\end{equation}
The loss is calculated as the cross-entropy between the task label $y_{c}$ and the softmax of the score $s$ normalized over the scores for all label verbalizations~\cite{schick2020exploiting}, weighted by a term $\alpha_{c}$.
\begin{equation}
    q_{\mathbf{p}_{c}^{i}} = \frac{e^{s_{\mathbf{p}_{c}^{i}}(\cdot|\mathbf{x}_{c})}}{\sum_{l\in\mathcal{L}_{c}}e^{s_{\mathbf{p}_{c}^{i}}(l|\mathbf{x}_{c})}}
\end{equation}
\begin{equation}
    L_{c} = \alpha_{c} * \frac{1}{N}\sum_{n} H(y_{c}^{(n)},q^{(n)}_{\mathbf{p}_{c}^{i}})
\end{equation}
where $N$ is the batch size, $n$ is a sample in the batch, $H$ is the cross-entropy, and $\alpha_{c}$ is a hyperparameter weight given to task $c$. 

\METHOD~then proceeds in the same fashion as standard PET. Different models are trained for each PVP tuple in $\mathcal{P}$, and each model produces raw scores $s_{\mathbf{p}_{m}^{i}}$ for all samples in the unlabeled data. The final score for a sample is then a weighted combination of the scores of individual models.
\begin{equation}
    s(l|\mathbf{x}_{u}^{j}) = \sum_{i}w_{i}*s_{\mathbf{p}_{m}^{i}}(l|\mathbf{x}_{u}^{j})
\end{equation}
where the weights $w_{i}$ are calculated as the accuracy of model $\mathcal{M}_{i}$ on the train set $D_{m}$ before training. The final classification model is then trained using KL-divergence between the predictions of the model and the scores $s$ as target logits.

\subsection{\METHOD~for Scientific Exaggeration}
\label{sec:mtpet_task_details}
We use \METHOD~to learn from data labeled for both of our formulations of the problem (\textbf{T1}, \textbf{T2}). In this, the first step is to define PVPs for exaggeration detection (\textbf{T1}) and claim strength prediction (\textbf{T2}).

To do this, we develop an initial set of PVPs and use PETAL~\cite{DBLP:conf/coling/SchickSS20} to automatically find verbalizers which adequately represent the labels for each task. We then update the patterns manually and re-run PETAL, iterating as such until we find a satisfactory combination of verbalizers and patterns which adequately reflect the task. Additionally, we ensure that the patterns between \textbf{T1} and \textbf{T2} are roughly equivalent. This yields 2 patterns for each task, provided in \autoref{tab:patterns}, and verbalizers given in \autoref{tab:verbalizers1}. The verbalizers found by PETAL capture multiple aspects of the task labels, selecting words such as ``mistaken,'' ``wrong,'' and ``artificial'' for exaggeration, ``preliminary'' and ``conditional'' for downplaying, and multiple levels of strength for strength detection such as ``estimated'' (correlational), ``cautious'' (conditional causal), and ``proven'' (direct causal).

For unlabeled data, we start with unlabeled pairs of full text press releases and abstracts. As we are concerned with detecting exaggeration in the primary conclusions, we first train a classifier based on single task PET for conclusion detection using a set of seed data. The patterns and verbalizers we use for conclusion detection are given in \autoref{tab:conc_patterns} and \autoref{tab:verbalizers2}. After training the conclusion detection model, we apply it to the press releases and abstracts, choosing the sentence from each with the maximum score $s_{\mathbf{p}}(1|\mathbf{x})$.
\begin{table}
    \def\arraystretch{1.1}
    \centering
    \fontsize{10}{10}\selectfont
    \begin{tabular}{c c p{3.5cm}}
    \toprule 
    Pattern & Label & Verbalizers\\
    \midrule 
        & Downplays& preliminary, competing, uncertainties\\
       $P_{T_{1}}^{0}$ & Same& following, explicit\\
        & Exaggerates& mistaken, wrong, hollow, naive, false, lies\\
    \midrule
        & Downplays& hypothetical, theoretical, conditional\\
       $P_{T_{1}}^{1}$ & Same& identical\\
        & Exaggerates& mistaken, wrong, premature, fantasy, noisy, artifical\\
    \midrule
        \multirow{12}{*}{$P_{T_{2}}^{*}$}& NA& sufficient, enough, authentic, medium\\
       & Correlational& inferred, estimated, calculated, borderline, approximately, variable, roughly\\
        & Cond. Causal& cautious, premature, uncertain, conflicting, limited\\
        & Causal& touted, proven, replicated, promoted, distorted\\
    \bottomrule 

    \end{tabular}
    \caption{Verbalizers for PVPs from both \textbf{T1} and \textbf{T2}. Verbalizers are obtained using PETAL~\cite{DBLP:conf/coling/SchickSS20}, starting with the top 10 verbalizers per label and then manually filtering out words which do not make sense with the given labels.}
    \vspace{-4mm}
    \label{tab:verbalizers1}
\end{table}

\section{Data Collection}
One of the main contributions of this work is a curated benchmark dataset for scientific exaggeration detection. Labeled datasets exist for the related task of claim strength detection in scientific abstracts and press releases~\cite{yu2020measuring,yu2019detecting}, but these 
data are from press releases and abstracts which are unrelated (i.e. the given press releases are not written about the given abstracts), making them unsuitable for benchmarking exaggeration detection. Given this, we curate a dataset of paired sentences from abstracts and associated press releases, labeled by experts for exaggeration based on their claim strength. We then collect a large set of unlabeled press release/abstract pairs useful for semi-supervised learning.

\subsection{Gold Data}
The gold test data used in this work are from~\citet{sumner2014association} and \citet{bratton2019association}, who annotate scientific papers, their abstracts, and associated press releases along several dimensions to characterize how press releases exaggerate papers. 
The original data consists of 823 pairs of abstracts and press releases. The 462 pairs from \citet{sumner2014association} have been used in previous work to test claim strength prediction~\cite{DBLP:conf/emnlp/LiZY17}, but the data, which contain press release and abstract conclusion sentences that are mostly paraphrases of the originals, are used as is.

We focus on the annotations provided for claim strength. The annotations consist of six labels which we map to the four labels defined in \citet{DBLP:conf/emnlp/LiZY17}. The labels and their meaning are given in \autoref{tab:all_labels}. 
This gives a claim strength label $l_{\rho}$ for the press release and $l_{\gamma}$ for the abstract. The final exaggeration label is then defined as follows:
\begin{equation*}
l_{e} = \begin{cases}
    0 & l_{\rho} < l_{\gamma}\\
    1 & l_{\rho} = l_{\gamma}\\
    2 & l_{\rho} > l_{\gamma}
    \end{cases}
\end{equation*}

\begin{table}[htp]
    \def\arraystretch{1.5}
        \centering
        \fontsize{10}{10}\selectfont
        \begin{tabular}{c c p{3.5cm}}
        \toprule 
        Name & Pattern\\
        \midrule 
    $P_{0}(x)$&\texttt{[MASK]}: $a$ \\
    $P_{1}(x)$&\texttt{[MASK]} - $a$\\
    $P_{2}(x)$&``\texttt{[MASK]}'' statement: $a$\\
    $P_{3}(x)$&$a$ (\texttt{[MASK]}) \\
    $P_{4}(x)$&(\texttt{[MASK]}) $a$\\
    $P_{5}(x)$&[Type: \texttt{[MASK]}] $a$ \\
\bottomrule
    \end{tabular}
    \caption{Patterns for conclusion detection.}
    \label{tab:conc_patterns}
\end{table}
\begin{table}[htp]
    \def\arraystretch{1.2}
    \centering
    \fontsize{10}{10}\selectfont
    \begin{tabular}{c l}
    \toprule 
    Label & Verbalizers\\
    \midrule 
        0& Text\\
        1& Conclusion\\
    \bottomrule 

    \end{tabular}
    \caption{Verbalizers for PVPs for conclusion detection.}
    \label{tab:verbalizers2}
\end{table}

As the original abstracts in the study are not provided, we automatically collect them using the Semantic Scholar API.\footnote{\url{https://api.semanticscholar.org/}}
We perform a manual inspection of abstracts to ensure the correct ones are collected, discarding missing and incorrect abstracts. 
Gold conclusion sentences are obtained by sentence tokenizing abstracts using SciSpaCy~\cite{neumann-etal-2019-scispacy} and finding the best matching sentence to the provided paraphrase in the data using ROUGE score~\cite{lin2004rouge}. We then manually fix sentences which do not correspond to a single sentence from the abstract. Gold press release sentences are gathered in the same way from the provided press releases.

This results in a dataset of 663 press release/abstract pairs labeled for claim strength and exaggeration. The label distribution is given in \autoref{tab:gold_statistics}. 
We randomly sample 100 of these instances as training data for few shot learning (\textbf{T1}), leaving 553 instances for testing. Additionally, we create a small training set of 1,138 sentences labeled for whether or not they are the main conclusion sentence of the press release or abstract. This data is used in the first step of \METHOD~to identify conclusion sentences in the unlabeled pairs.

\begin{table*}[t]
    \def\arraystretch{1.3}
    \centering
    \fontsize{10}{10}\selectfont
    \begin{tabular}{c p{4cm} c p{4cm}}
    \toprule 
    \citet{sumner2014association} & Description & \citet{DBLP:conf/emnlp/LiZY17}& Description\\
    \midrule 
    0& No relationship mentioned& - & -\\
    \hline
    1& Statement of no relationship & 0& Statement of no relationship\\
    \hline
    2& Statements of correlation & \multirow{2}{*}{1}& \multirow{2}{*}{Statement of correlation}\\
    3& Ambiguous statement of relationship & &\\
    \hline
    4& Conditional statement of causation &\multirow{2}{*}{2}& \multirow{2}{4cm}{Conditional statement of causation}\\
    5& Statement of ``can'' & &\\
    \hline
    6& Statements of causation & 3& Statement of causation\\
    \bottomrule 

    \end{tabular}
    \caption{Claim strength labels and their meaning from the original data in \citet{sumner2014association} and \citet{bratton2019association} and the mappings to the labels from \citet{DBLP:conf/emnlp/LiZY17}. We use the labels from \citet{DBLP:conf/emnlp/LiZY17} in this study, including for deriving the exaggeration labels.}
    \label{tab:all_labels}
\end{table*}
\begin{table}
    \centering
    \fontsize{10}{10}\selectfont
    \begin{tabular}{l c}
    \toprule 
     Label & Count\\
    \midrule
        Downplays & 113\\
        Same & 406\\
        Exaggerates& 144\\
    \bottomrule 

    \end{tabular}
    \caption{Number of labels per class for benchmark exaggeration detection data.}
    \label{tab:gold_statistics}
\end{table}
\begin{table}
    \def\arraystretch{1.2}
    \centering
    \fontsize{10}{10}\selectfont
    \begin{tabular}{l c c c}
    \toprule 
    Method & P & R & \multicolumn{1}{c}{F1} \\
    \midrule 
       Supervised & $28.06$& $33.10$& $29.05$\\
       PET & $41.90$& $39.87$& $39.12$\\
       MT-PET & $\mathbf{47.80}$& $\mathbf{47.99}$& $\mathbf{47.35}$\\
    \bottomrule 

    \end{tabular}
    \caption{Results for exaggeration detection with paired conclusion sentences from abstracts and press releases (\textbf{T1}). MT-PET uses 200 sentences for strength classification, 100 each from press releases and abstracts.} 
    \label{tab:t1_base_results}
\end{table}
\begin{table*}
    \def\arraystretch{1.2}
    \centering
    \fontsize{10}{10}\selectfont
    \begin{tabular}{l c c c c c c}
    \toprule 
    Method & $|$\textbf{T2}$|$,$|$\textbf{T1}$|$& P & R & \multicolumn{1}{c}{F1} & Press F1 & Abstract F1 \\
    \midrule 
       Supervised & 200,0 & $49.28$& $51.07$& $49.03$& $54.78$& $59.41$\\
       PET& 200,0 & $55.76$& $58.58$& $56.57$& $63.56$& $62.76$\\
       MT-PET& 200,100 & $\mathbf{56.68}$& $\mathbf{60.13}$& $\mathbf{57.44}$& $\mathbf{64.72}$& $\mathbf{63.27}$\\
     \midrule 
       Supervised& 4500,0 & $58.20$& $59.99$& $58.66$& $63.26$& $\textbf{67.26}$\\
       PET& 4500,0 & $59.53$& $61.84$& $60.45$& $\mathbf{64.20}$& $64.92$\\
       MT-PET& 4500,100 & $\mathbf{60.09}$& $\mathbf{62.68}$& $\mathbf{61.11}$& $63.93$& $64.69$\\
    \midrule
       PET+in domain MLM& 200,100 & $\textit{57.18}$& $\textit{60.12}$& $\textit{58.06}$& $\textit{64.29}$& $\textit{62.69}$\\
       PET+in domain MLM& 4500,100 & $\textit{59.87}$& $\textit{62.33}$& $\textit{60.85}$& $\textit{64.10}$& $\textit{64.73}$\\
    \bottomrule 

    \end{tabular}
    \caption{Results on exaggeration detection via strength classification (\textbf{T2}) with varying numbers of instances. \METHOD~uses 100 instances from paired press and abstract sentences (200 sentences total).} 
    \label{tab:t2_base_results}
\end{table*}

For \textbf{T2} we use the data from~\citet{yu2020measuring,yu2019detecting}. \citet{yu2019detecting} create a dataset of 3,061 conclusion sentences labeled for claim strength from structured PubMed abstracts of health observational studies with conclusion sections of 3 sentences or less. \citet{yu2020measuring} then annotate statements from press releases from EurekAlert. The selected data are from the title and first two sentences of the press releases, as \citet{sumner2014association} note that most press releases contain their main conclusion statements in these sentences, following an inverted pyramid structure common in journalism~\cite{po2003news}. Both studies use the labeling scheme from \citet{DBLP:conf/emnlp/LiZY17} (see \autoref{tab:all_labels}). 
The final data contains 2,076 labeled conclusion statements. From these two datasets, we select a random stratified sample of 4,500 instances for training in our full-data experiments, and subsample 200 for few-shot learning (100 from abstracts and 100 from press releases). 

\subsection{Unlabeled Data}
We collect unlabeled data from ScienceDaily,\footnote{\url{https://www.sciencedaily.com/}} a science reporting website which aggregates and re-releases press releases from a variety of sources. To do this, we crawl press releases from ScienceDaily via the Internet Archive Wayback Machine\footnote{\url{https://archive.org/web/}} between January 1st 2016 and January 1st 2020 using Scrapy.\footnote{\url{https://scrapy.org/}} We discard press releases without paper DOIs and then pair each press release with a paper abstract by querying for each DOI using the Semantic Scholar API. This results in an unlabeled set of 7,741 press release/abstract pairs. Additionally, we use only the title, lead sentence, and first three sentences of each press release.

\section{Experiments}
Our experiments are focused on the following primary research questions:
\begin{itemize}[noitemsep]
    \item \textbf{RQ1}: Does \METHOD~improve over PET for scientific exaggeration detection?
    \item \textbf{RQ2}: Which formulation of the problem leads to the best performance?
    \item \textbf{RQ3}: Does few-shot learning performance approach the performance of models trained with many instances?
    \item \textbf{RQ4}: What are the challenges of scientific exaggeration prediction?
\end{itemize}
We experiment with the following model variants:
\begin{itemize}[noitemsep]
    \item \textbf{Supervised}: A fully supervised setting where only labeled data is used.
    \item \textbf{PET}: Standard single-task PET.
    \item \textbf{\METHOD}: We run \METHOD~with data from one task formulation as the main task and the other formulation as the auxiliary task.
\end{itemize}

We perform two evaluations in this setup: one with \textbf{T1} as the main task and one with \textbf{T2}. For \textbf{T1}, we use the 100 expert annotated instances with paired press release and abstract sentences labeled for exaggeration (200 sentences total). For \textbf{T2}, we use 100 sentences from the press data from \citet{yu2020measuring} and 100 sentences from the abstract data in \citet{yu2019detecting} labeled for claim strength. We use RoBERTa base~\cite{DBLP:journals/corr/abs-1907-11692} from the HuggingFace Transformers library~\cite{DBLP:conf/emnlp/WolfDSCDMCRLFDS20} as the main model, and set $\alpha_{m}$ to be $1$, and $\alpha_{a} = \text{min}(2, \frac{|D_{m}|}{|D_{a}|})$. All methods are evaluated using macro-F1 score, and results are reported as the average performance over 5 random seeds.

\subsection{Performance Evaluation}
We first examine the performance with \textbf{T1} as the base task (see \autoref{tab:t1_base_results}). In a purely supervised setting, the model struggles to learn and mostly predicts the majority class. Basic PET yields a substantial improvement of 10 F1 points, with \METHOD~further improving upon this by another 8 F1 points. Accordingly, we conclude that training with auxiliary task data provides much benefit for scientific exaggeration detection in the \textbf{T1} formulation.

We next examine performance with \textbf{T2} (strength classification) as the main task in both few-shot and full data settings (see \autoref{tab:t2_base_results}). In terms of base performance, the model can predict exaggeration better than \textbf{T1 }in a purely supervised setting. For PET and \METHOD, we see a similar trend; with 200 instances for \textbf{T2}, PET improves by 7 F1 points over supervised learning, and \METHOD~improves on this by a further 0.9 F1 points. Additionally, \METHOD~improves performance on the individual tasks of predicting the claim strength of conclusions in press releases and scientific abstracts with 200 examples. While less dramatic, we still see gains in performance using PET and \METHOD~when 4,500 instances from \textbf{T2} are used, despite the fact that there are still only 100 instances from \textbf{T1}. We also test if the improvement in performance is simply due to training on more in-domain data (``PET+in domain MLM'' in \autoref{tab:t2_base_results}). We observe gains for exaggeration detection using masked language modeling on data from \textbf{T1}, but \METHOD~still performs better at classifying the strength of claims in press releases and abstracts when 200 training instances from \textbf{T2} are used.

\paragraph{RQ1} Our results indicate that \METHOD~does in fact improve over PET for both training setups. With \textbf{T1} as the main task and \textbf{T2} as the auxiliary task, we see that performance is substantially improved, demonstrating that learning claim strength prediction helps produce soft-labeled training data for \textit{exaggeration detection}. Additionally, we find that the reverse holds with \textbf{T2} as main task and \textbf{T1} as auxiliary task. As performance can also be improved via masked language modeling on data from \textbf{T1}, this indicates that some of the performance improvement could be due to including data closer to the test domain. However, our error analysis in \autoref{sec:error_analysis} shows that these methods improve model performance on different types of data.

\paragraph{RQ2} We find that \textbf{T2} is better suited for scientific exaggeration detection in this setting, however, with a couple of caveats. First, the final exaggeration label is based on expert annotations for claim strength, so clearly claim strength prediction will be useful in this setup. Additionally, the task may be more forgiving here, as only the direction needs to be correct and not necessarily the final strength label (i.e. predicting `0' for the abstract and any of `1,' `2,' or `3' for the press release label will result in an exaggeration label of `exaggerates').
\begin{figure}[t]
     \centering
         \includegraphics[width=\linewidth]{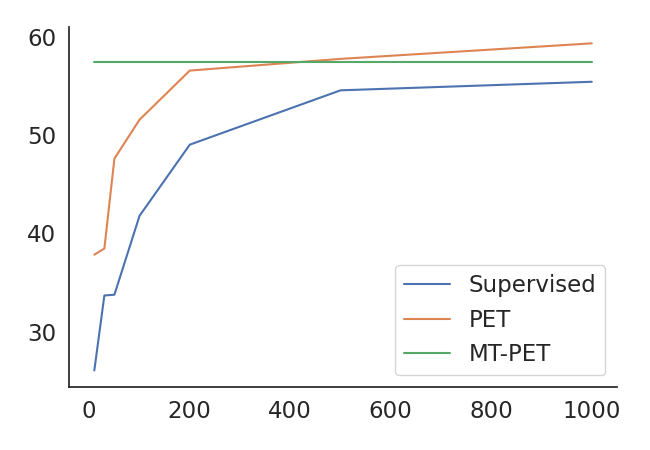}
         \caption{Learning curve for supervised learning and PET compared to performance of \METHOD~using 200 instances from \textbf{T2} and 100 from \textbf{T1}.}
         \label{fig:learning_curve}
\end{figure}
\paragraph{RQ3} We next examine the learning dynamics of our few-shot models with different amounts of training data (see \autoref{fig:learning_curve}), comparing them to \METHOD~to understand how well it performs compared to settings with more data. \METHOD~ with only 200 samples is highly competitive with purely supervised learning on 4,500 samples (57.44 vs. 58.66). Additionally, \METHOD~performs at or above supervised performance up to 1000 input samples, and at or above PET up to 500 samples, again using only 200 samples from \textbf{T2} and 100 from \textbf{T1}.

\subsection{Error Analysis}
\label{sec:error_analysis}
\begin{figure*}
     \centering

         \centering
         \includegraphics[width=0.95\textwidth]{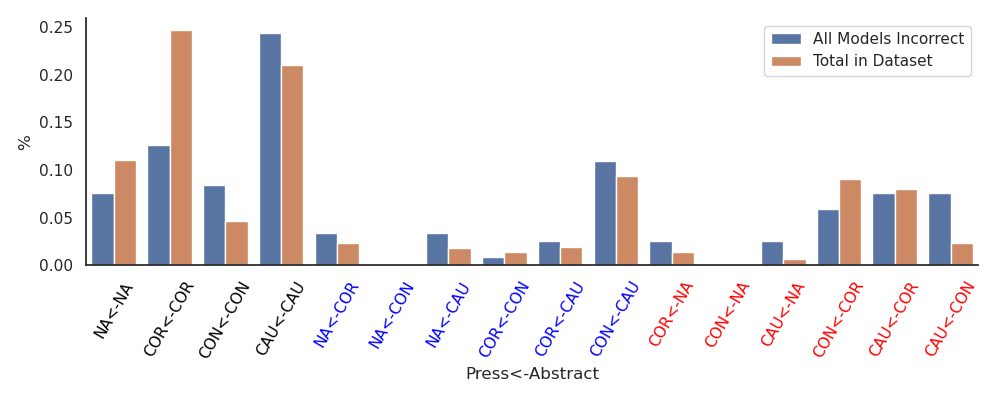}
         \caption{Proportion of examples by label which all models predict incorrectly.}
         \label{fig:all_fail}
\end{figure*}
\paragraph{RQ4} Finally, we try to understand the difficulty of scientific exaggeration detection by observing where models succeed and fail (see \autoref{fig:all_fail}). The most difficult category of examples to predict involve direct causal claims, particularly exaggeration and downplaying when one document is a direct causal claim and the other an indirect causal claim (`CON->CAU', `CAU->CON'). Also, it is challenging to predict when both the press release and abstract conclusions are directly causal.

The models have the easiest time predicting when both statements involve correlational claims, and exaggerations involving correlational claims from abstracts. We also observe that \METHOD~helps the most for the most difficult category: causal claims (see \autoref{fig:supervised_succeed} in \autoref{sec:extra_plots}). The model is particularly better at differentiating when a causal claim in an abstract is \textit{downplayed} by a press release. It is also better at identifying correlational claims than PET, where many claims involve association statements such as `linked to,' `could predict,' `more likely,` and `suggestive of.'

The model trained with MLM on data from \textbf{T1} also benefits causal statement prediction, but mostly for when both statements are causal, whereas \METHOD~sees more improvement for pairs where one causal statement is exaggerated or downplayed by another (see \autoref{fig:pet_succeed} in \autoref{sec:extra_plots}). This suggests that training with the patterns from \textbf{T1} helps the model to differentiate direct causal claims from weaker claims, while MLM training mostly helps the model to understand better how direct causal claims are written. We hypothesize that combining the two methods would lead to mutual gains.

\section{Related Work}
\subsection{Scientific Misinformation Detection}
Misinformation detection focuses on a variety of problems, including fact verification~\cite{DBLP:conf/naacl/ThorneVCM18,augenstein-etal-2019-multifc}
, check-worthiness detection~\cite{wright2020fact,DBLP:conf/ecir/NakovMEBMSAHHBN21}
, stance~\cite{augenstein-etal-2016-stance,baly-etal-2018-integrating,hardalov2021survey} and clickbait detection~\cite{potthast-etal-2018-crowdsourcing}. While most work has focused on social media and general domain text, recent work has begun to explore different problems in detecting misinformation in scientific text such as SciFact~\cite{DBLP:conf/emnlp/WaddenLLWZCH20} and CiteWorth~\cite{wright2021citeworth}, as well as related tasks such as summarization \cite{DBLP:journals/corr/abs-2104-06486,dangovski2021we}.

Most work on scientific exaggeration detection has focused on flagging when the primary finding of a scientific paper has been exaggerated by a press release or news article~\cite{sumner2014association,bratton2019association,yu2020measuring,yu2019detecting,DBLP:conf/emnlp/LiZY17}. 
\citet{sumner2014association} and \citet{bratton2019association} manually label pairs of press releases and scientific papers on a wide variety of metrics, finding that one third of press releases contain exaggerated claims, and 40\% contain exaggerated advice. 
\citet{DBLP:conf/emnlp/LiZY17} is the first study into automatically predicting claim strength, using the data from \citet{sumner2014association} as a small labeled dataset. \citet{yu2019detecting} and \citet{yu2020measuring} extend this by building larger datasets for claim strength prediction, performing an analysis of a large set of unlabeled data to estimate the prevalence of claim exaggeration in press releases. Our work improves upon this by providing a more realistic task formulation of the problem, consisting of paired press releases and abstracts, as well as curating both labeled and unlabeled data to evaluate methods in this setting. 

\subsection{Learning from Task Descriptions}
Using natural language to perform zero and few-shot learning has been demonstrated on a number of tasks, including question answering~\cite{radford2018improving}, text classification~\cite{DBLP:journals/corr/abs-1912-10165}, relation extraction~\cite{DBLP:conf/aaai/BouraouiCS20} and stance detection \cite{hardalov2021emnlp,hardalov2021fewshot}. Methods of learning from task descriptions have been gaining more popularity since the creation of GPT-3~\cite{DBLP:conf/nips/BrownMRSKDNSSAA20}. 
\citet{DBLP:journals/jmlr/RaffelSRLNMZLL20} attempt to perform this with smaller language models by converting tasks into natural language and predicting tokens in the vocabulary. \citet{schick2020exploiting} propose PET, a method for few shot learning which converts tasks into cloze-style QA problems which can be solved by a pretrained language model in order to provide soft-labels for unlabeled data. We build on PET, showing that complementary cloze-style QA tasks can be trained on simultaneously to improve few-shot performance on scientific exaggeration detection.

\section{Conclusion}
In this work, we present a formalization of and investigation into the problem of scientific exaggeration detection. As data for this task is limited, we develop a gold test set for the problem and propose \METHOD, a semi-supervised approach based on PET, to solve it with limited training data. We find that \METHOD~helps in the more difficult cases of identifying and differentiating direct causal claims from weaker claims, and that the most performant approach involves classifying and comparing the individual claim strength of statements from the source and target documents. The code and data for our experiments can be found online\footnote{\url{https://github.com/copenlu/scientific-exaggeration-detection}}. Future work should focus on building more resources e.g. datasets for exploring scientific exaggeration detection, including data from multiple domains beyond health science. 
Finally, it would be interesting to explore how MT-PET works on combinations of more general NLP tasks, such as question answering and natural language inference or part-of-speech tagging and named entity recognition.


\section*{Acknowledgements}

$\begin{array}{l}\includegraphics[width=1cm]{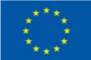} \end{array}$ The research documented in this paper has received funding from the European Union's Horizon 2020 research and innovation programme under the Marie Sk\l{}odowska-Curie grant agreement No 801199.

\section*{Broader Impact Statement}

Being able to automatically detect whether a press release exaggerates the findings of a scientific article could help journalists write press releases, which are more faithful to the scientific articles they are describing. We further believe it could benefit the research community working on fact checking and related tasks, as developing methods to detect subtle differences in a statement's veracity is currently understudied.

On the other hand, as our paper shows, this is currently still a very challenging task, and thus, the resulting models should only be applied in practice with caution. 
Moreover, it should be noted that the predictive performance results reported in this paper are for press releases written by science journalists -- one could expect worse results for press releases which more strongly simplify scientific articles.

\bibliography{anthology,custom}
\bibliographystyle{acl_natbib}

\newpage
\appendix

\section{Error Analysis Plots}
\label{sec:extra_plots}
\begin{figure*}[t]
     \centering

         \centering
         \includegraphics[width=0.8\textwidth]{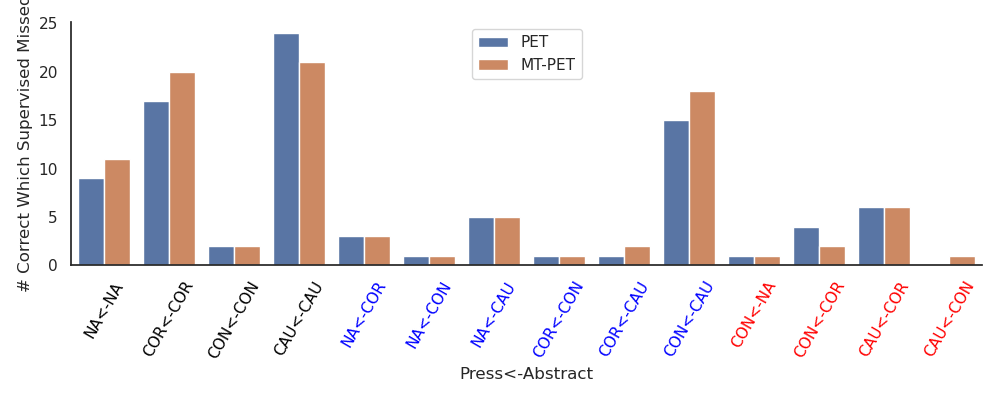}
         \caption{Number of instances that each model predicted correctly which the supervised model predicted incorrectly.}
         \label{fig:supervised_succeed}
\end{figure*}

\begin{figure*}[t]
     \centering

         \centering
         \includegraphics[width=0.8\textwidth]{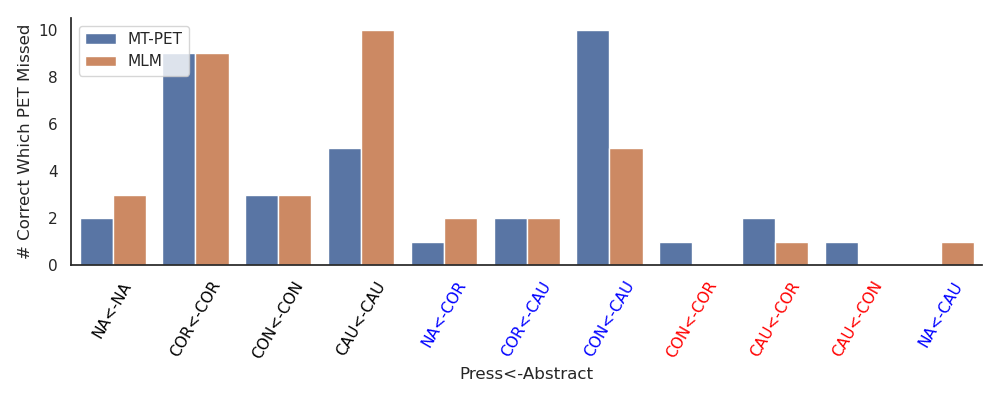}
         \caption{Number of instances that each model predicted correctly which PET predicted incorrectly.}
         \label{fig:pet_succeed}
\end{figure*}

Extra plots from our error analysis are given in \autoref{fig:supervised_succeed} and \autoref{fig:pet_succeed}.

\section{Reproducibility}
\subsection{Computing Infrastructure}

All experiments were run on a shared cluster. Requested jobs consisted of 16GB of RAM and 4 Intel Xeon Silver 4110 CPUs. We used a single NVIDIA Titan X GPU with 24GB of RAM.

\subsection{Average Runtimes}
The average runtime performance of each model is given in \autoref{tab:runtimes}. Note that different runs may have been placed on different nodes within a shared cluster. 
\begin{table}[htp]
    \centering
    \fontsize{10}{10}\selectfont
    \begin{tabular}{l c c}
    \toprule 
     Setting & |\textbf{T1}|,|\textbf{T2}| & Time\\
    \midrule
        Supervised& 100,0 & 00h01m28s\\
        PET& 100,0 & 00h11m14s\\
        \METHOD& 100,200 & 00h13m05s\\
        \midrule
        Supervised& 0,200 & 00h01m20s\\
        PET& 0,200 & 00h16m22s\\
        \METHOD& 100,200 & 00h18m43s\\
        Supervised& 0,4500 & 00h03m23s\\
        PET& 0,4500 & 00h40m23s\\
        \METHOD& 100,4500& 00h31m48s\\
    \bottomrule 

    \end{tabular}
    \caption{Average runtimes for each model (runtimes are taken for the entire run of an experiment).}
    \label{tab:runtimes}
\end{table}

\subsection{Number of Parameters per Model}
We use RoBERTa, specifically the base model, for all experiments, which consists of 124,647,170 parameters.

\subsection{Validation Performance}
As we are testing a few shot setting, we do not use a validation set and only report the final test results.

\subsection{Evaluation Metrics}
The primary evaluation metric used was macro F1 score.
We used the sklearn implementation of \texttt{precision\_recall\_fscore\_support} for F1 score, which can be found here: \url{https://scikit-learn.org/stable/modules/generated/sklearn.metrics.precision_recall_fscore_support.html}. Briefly:
\begin{equation*}
   p = \frac{tp}{tp + fp} 
\end{equation*}
\begin{equation*}
   r = \frac{tp}{tp + fn} 
\end{equation*}
\begin{equation*}
   F1 = \frac{2*p*r}{p+r} 
\end{equation*}
where $tp$ are true positives, $fp$ are false positives, and $fn$ are false negatives. Macro F1 is average F1 across all classes.

\subsection{Hyperparameters}
\label{sec:hyperparams}
\paragraph{T1 Hyperparameters Supervised/PET training} We used the following hyperparameters for experiments with \textbf{T1} as the main task: Epochs: 10; Batch Size: 4; Learning Rate: 0.00005598; Warmup Steps: 50; Weight decay: 0.001. We also weigh the cross-entropy loss based on the label distribution. These hyperparameters are found by performing a hyperparameter search using 4-fold cross validation on the 100 training examples. The bounds are as follows: Learning rate: $[0.000001, 0.0001$; Warmup steps: $\{0, 10, 20, 30, 40, 50, 100\}$; Batch size: $\{4, 8\}$; Weight decay: $\{0.0, 0.0001, 0.001, 0.01, 0.1\}$; Epochs: $[2, 10]$.

\paragraph{T2 Hyperparameters Supervised/PET training} Epochs: 10; Batch Size: 4; Learning Rate: 0.00003; Warmup Steps: 50; Weight Decay: 0.001. We also weigh the cross-entropy loss based on the label distribution.

\paragraph{Hyperparameters for Distillation} We used the following hyperparameters for distillation (training the final classifier after PET) for both \textbf{T1} and \textbf{T2} as the main task: Epochs: 3; Batch Size: 4; Learning Rate: 0.00001; Warmup Steps: 200; Weight decay: 0.01; Temperature: 2.0. We also weigh the cross-entropy loss based on the label distribution.

\subsection{Data}
\label{sec:data_info}
We build our benchmark test dataset from the studies of \citet{sumner2014association} and \citet{bratton2019association}. The original data can be found at \url{https://figshare.com/articles/dataset/InSciOut/903704} and \url{https://osf.io/v8qhn/files/}. A link to the test data will be provided upon acceptance of the paper (and included in the supplemental material). Claim strength data from \citet{yu2019detecting} for abstracts can be found at \url{https://github.com/junwang4/correlation-to-causation-exaggeration/blob/master/data/annotated_pubmed.csv}. Claim strength data for press releases from \citet{yu2020measuring} can be found at \url{https://github.com/junwang4/correlation-to-causation-exaggeration/blob/master/data/annotated_eureka.csv} 



\end{document}